\documentclass[conference]{IEEEtran}
\IEEEoverridecommandlockouts
\usepackage{etoolbox}
\makeatletter
\patchcmd{\@makecaption}
  {\scshape}
  {}
  {}
  {}
\makeatother
\usepackage{threeparttable, booktabs}
\usepackage{cite}
\usepackage{amsmath,amssymb,amsfonts}
\usepackage{algorithmic}
\usepackage{algorithm}
\usepackage{graphicx}
\usepackage{booktabs}
\usepackage{textcomp}
\usepackage{hyperref}
\def\BibTeX{{\rm B\kern-.05em{\sc i\kern-.025em b}\kern-.08em
    T\kern-.1667em\lower.7ex\hbox{E}\kern-.125emX}}
\begin{document}

\title{Estimating the effective dimension of large biological datasets using Fisher separability analysis\\
\thanks{\IEEEauthorrefmark{1} These Authors contributed equally to this work. \IEEEauthorrefmark{2} To whom correspondence should be addressed. This project was supported by the Ministry of education and science of Russia (Project No. 14.Y26.31.0022)}
}

\author{\IEEEauthorblockN{Luca Albergante\IEEEauthorrefmark{1}}
\IEEEauthorblockA{
\textit{Institut Curie, PSL Research University,}\\
\textit{Mines ParisTech, INSERM, U900}\\
Paris, France\\
luca.albergante@curie.fr}
\and
\IEEEauthorblockN{Jonathan Bac\IEEEauthorrefmark{1}}
\IEEEauthorblockA{
\textit{Institut Curie, PSL Research University,}\\
\textit{Mines ParisTech, INSERM, U900}\\
\textit{Centre de Recherches Interdisciplinaires,}\\
\textit{Paris Diderot University}\\
Paris, France\\
jonathan.bac@cri-paris.org}
\and
\IEEEauthorblockN{Andrei Zinovyev\IEEEauthorrefmark{2}}
\IEEEauthorblockA{
\textit{Institut Curie, PSL Research University,}\\
\textit{Mines ParisTech, INSERM, U900}\\
Paris, France \\
\textit{Lobachevsky University}
\\
Nizhni Novgorod, Russia \\
andrei.zinovyev@curie.fr} \\

}

\maketitle

\begin{abstract}
Modern large-scale datasets are frequently said to be high-dimensional. However, their data point clouds frequently possess structures, significantly decreasing their intrinsic dimensionality (ID) due to the presence of clusters, points being located close to low-dimensional varieties or fine-grained lumping. We test a recently introduced dimensionality estimator, based on analysing the separability properties of data points, on several benchmarks and real biological datasets. We show that the introduced measure of ID has performance competitive with state-of-the-art measures, being efficient across a wide range of dimensions and performing better in the case of noisy samples. Moreover, it allows estimating the intrinsic dimension in situations where the intrinsic manifold assumption is not valid.
\end{abstract}

\begin{IEEEkeywords}
high-dimensional data, data intrinsic dimension, intrinsic dimensionality, separability, biological datasets, single cell RNA-Seq
\end{IEEEkeywords}

\section{Introduction}

\begin{figure}[ht]
{\centering\includegraphics[width=8cm]{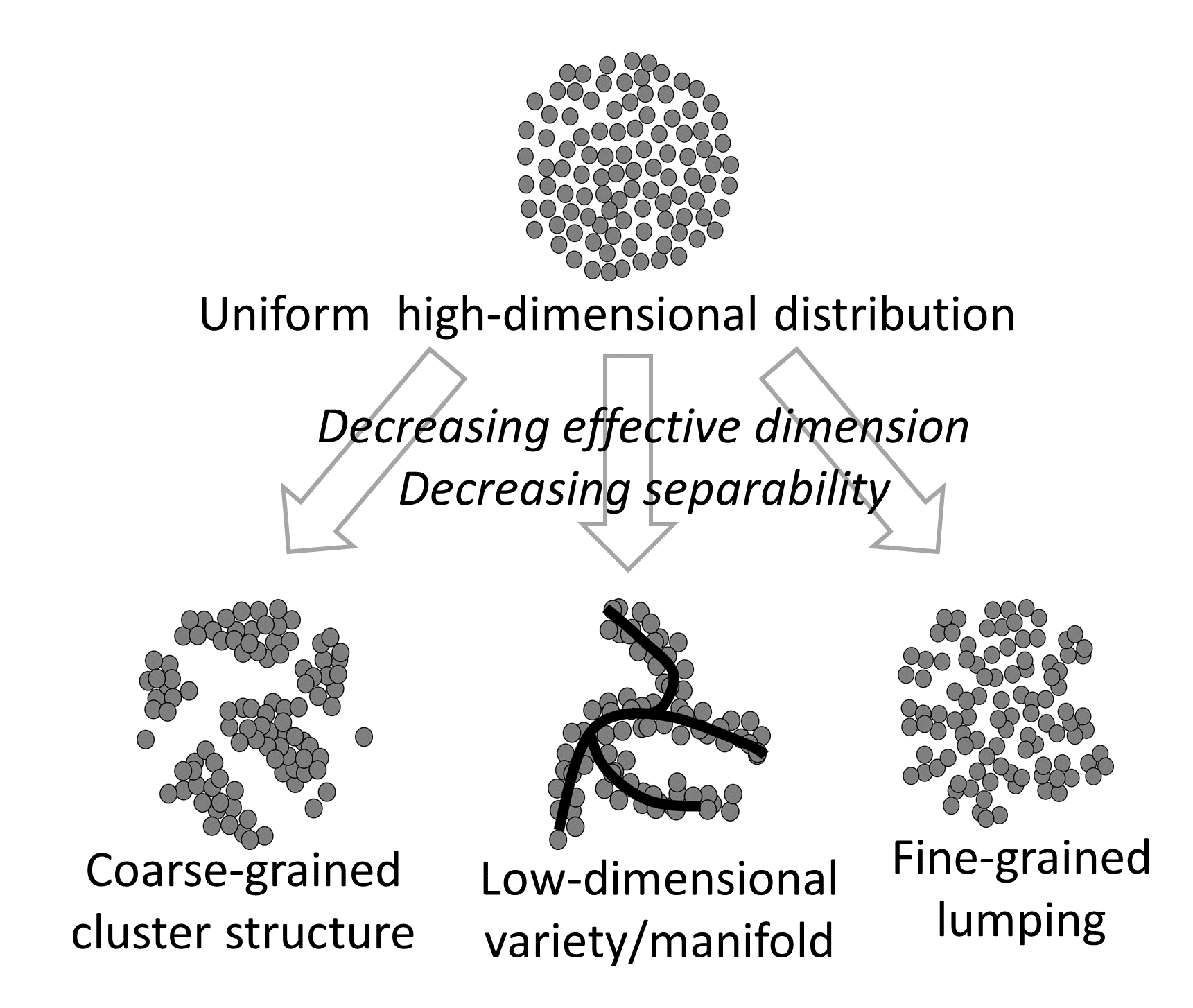}}
\caption{Stereotypical scenarios of real-life high-dimensional data point cloud organisation, affecting the effective dimensionality of the data.
(Top) Data point cloud can be modelled as a close-to-uniform multidimensional distribution (and benefit from ``blessing of dimensionality" in high dimensions).  (Bottom-right) Data point cloud can be organized into a relatively small number of well-defined clusters. (Bottom-middle) Data point clouds can be located close to a low-dimensional variety (or manifold, in simple cases). (Bottom-left) Data point cloud can be characterised by a fine-grained lumping (heterogeneity) which can not be well represented as being located close to a manifold of low dimension. In cases represented in the top-right and bottom panels the data cloud is characterised by lower ID than the full (ambient) dimension of the data space. The figure is drawn from personal communication with A.N.Gorban and I.Tyukin who coined the term ``fine-grained lumping".
\label{Figure1}}.
\end{figure}

High-dimensional data are becoming increasingly available in real-life problems across many disciplines. Multiple research efforts in the field of machine learning are currently focused on better characterising, analysing, and comprehending them. A key feature related to data complexity, which is still largely unexplored, is the \textit{intrinsic dimensionality (ID)}, sometimes also called effective dimensionality, of the cloud of points. Informally, ID describes the effective number of variables needed to approximate the data with sufficient accuracy. ID can be measured both globally and locally (i.e., by segmenting the data cloud)\cite{Campadelli2015}.

Different approaches have been used to formalise the concept of ID. We refer the reader to other works for a list and comparisons of the different definitions and a discussion of their properties\cite{Campadelli2015,Camastra2016}. A compact presentation of the currently used definitions is also available in Section \ref{JonRev}. 

Despite the great diversity of currently used approaches for defining the intrinsic data dimensionality, most of them assume that there exists a relatively low-dimensional variety  embedded into the high-dimensional space around which the data cloud is organised. Moreover, it is frequently assumed that the nature of this variety is a manifold, and that the data point cloud represents an i.i.d. sample from the manifold with some simple model of noise. In practice, the ID of the manifold is assumed to be not only much smaller than the number of variables defining the data space but also to be small in absolute number. Thus, any practically useful non-linear data manifold should not have more than three or four intrinsic degrees of freedom. 

Theoretically, the manifold concept does not have to be universal in the case of real-life datasets. Even if a low-dimensional variety exists, it can be more complex than a simple manifold: for example, it can contain branching points or be of variable local intrinsic dimension. Principal trees, graphs and principal cubic complexes (direct product of principal graphs as factors) have been suggested as a constructive approach to deal with such complex cases \cite{GORBAN2007382,Gorban2009}. In the case of existence of a well-defined cluster structure in the data cloud, the underlying variety can be thought of as discontinuous (e.g., the model of principal forest \cite{elpigraph}).  

Conversely, a typical mental image of a ``genuinely high-dimensional" data point cloud is a uniformly sampled $n$-dimensional sphere or a $n$-hypercube, where $n \gg 1$ (at least several tens). Interestingly, in this model, the data point cloud can enjoy the ``blessing of dimensionality", which results in any two data vectors being almost orthogonal and any data point being linearly separable from the rest of the data point cloud \cite{Gorban2018Blessing}. The separability properties can be used, for example, to provide simple non-destructive (not requiring retraining) correctors for the legacy AI systems \cite{GORBAN2018303,Gorban2017Theorems,Tykin2018Front}. In this sense, truly high-dimensional data distributions are characterised by surprising ``simplicity" as opposite to low-dimensional varieties which can possess rather complex non-linear branching or looping structure. 

However, real life datasets can be characterised by properties which are difficult to fit into such simple paradigms. In particular, real-life datasets are expected to significantly violate the i.i.d. sampling assumption. The data inhomogeneity can manifest by the existence of micro-clusters which are not globally organized into a low-dimensional structure \cite{GORBAN2018303}. These micro-clusters might be undetectable by standard clustering algorithms because of their small size, fuzziness and instability. Existence of such {\it fine-grained lumping} in the data (Figure~\ref{Figure1}) can be also thought of as a decrease in ID. For example, it leads to destroying the separability and measure concentration properties, making the data more similar to lower-dimensional (but uniformly distributed) data point clouds. 

A single data point cloud can combine regions with several structure types described above in different regions of the data space, and hence be characterised by variable ID. In this context, two important questions are: 1) what parts of the point cloud can be reasonably approximated by a low-dimensional object (e.g., principal curves or trees) and which can not? 2) for those parts which can not be described by a locally small intrinsic dimension, can we estimate how close we are to the ``blessing of dimensionality" scenario and can we profit from it, or not? In this work, we suggest an approach for answering these questions.

In recent works by A. Gorban, I. Tyukin and colleagues, the authors proved a series of stochastic separation theorems that can be used to define the properties of high-dimensional data distributions in an efficient and scalable fashion. The authors exploited a convenient framework of Fisher linear discriminants \cite{GORBAN2018303,Gorban2017Theorems,Gorban2018Blessing}. We show that such framework can be adapted to construct computationally efficient estimators of local dimensionality tackling different data organization types (Figure~\ref{Figure1}). This analysis will then be applied to biological data to show the value of dimensionality analysis in deriving actionable information from data.

Throughout the text, $R^n$ will denote the Euclidean $n$-dimensional linear real vector space, $\mathbf{x}= (x_1,\ldots,x_n)$ the elements of $R^n$, $(\mathbf{x},\mathbf{y}) = \sum_{k=1 \ldots n} x_k y_k$ the inner product of $\mathbf{x}$ and $\mathbf{y}$, and  $||\mathbf{x}||=(\mathbf{x},\mathbf{x})$ the standard Euclidean norm in $R^n$. $S^n \subset R^n$ will denote the unit $n$-dimensional sphere and $|Y|$ or $N$ the number of points in a finite set $Y$.

\section{Defining and measuring intrinsic dimensionality}\label{JonRev}

Despite being used in machine learning research, the term intrinsic dimensionality lacks a unique consensus definition\cite{Campadelli2015}. One of its first use traces back to the context of signal analysis, referring to the minimum number of parameters required by a signal generator to closely approximate each signal in a collection \cite{Bennett}. Other authors  \cite{bishop1995neural,Fuku1982} define the ID of a dataset to be $n$ if it lies entirely within a $n$-dimensional manifold in $R^n$ with none or little information loss. By shifting the attention from a finite set of points to a generating process, other authors say that the data generating process $Y_i$ has ID $n$ if $Y_i$ can be written as $Y_i = X_i + \epsilon_i$, where $X_i$ is sampled according to a probability measure with a smooth density and with a support on a smooth $n-$dimensional manifold, $\epsilon_i$ is a noise component which is small on a scale where $M$ is well approximated by a $n$-dimensional subspace \cite{Kerstin}. These definitions are grounded in the so-called \textit{Manifold hypothesis}, i.e. that data is sampled from an underlying $n$-dimensional manifold. Following this hypothesis, the goal of  ID estimation is to recover $n$.

While these definitions are very important to better comprehend the problem at hand, they do not provide a way to directly estimate ID. Over the years, researchers devised different estimators, which, however, can be roughly classified by their mode of operation (see \cite{Campadelli2015} for the details of these categories).

\textit{Topological methods} explicitly seek to estimate the topological dimension (e.g. as defined by the covering dimension) of a manifold. However, they are unsuitable for most practical applications \cite{Campadelli2015,mordohai2010,li2009}. \textit{Fractal methods} are grounded in the theory of fractal geometry and have been developed by adapting the ideas originally used to study strange attractors' dimensionality. \textit{Projective methods} use different approaches (such as MDS or PCA) that perform a mapping of the points into a relatively low-dimensional subspace, by minimising some cost function, which should not exceed certain threshold (e.g. the reconstruction error in ISOMAP) \cite{shepard1972multidimensional,jolliffe2002principal,Tenenbaum2319}. \textit{Graph-based methods} exploit scaling properties of graphs, such as the length of the geodesic minimum spanning tree \cite{Costa}. Finally, the \textit{Nearest neighbours} category includes those methods that work at the local level, and rely on distribution properties of local distances or angles. It is worth noting that some recent estimators in this category have been expressly designed to exploit properties of concentration of measure \cite{DANCo,Kerstin,ANOVA,SSV}. 

One of the most used in practice dimensionality estimator uses the fractal correlation dimension \cite{Grassberger1983}, which is based on the fact - also used by many other estimators - that the number of points contained in a ball of growing radius $r$ will scale exponentially with the dimension of the underlying $n$-manifold. This counting process is performed using the correlation sum : 
\begin{equation*}\label{corrdim}
C(r)=\lim_{N\rightarrow \infty}\frac{2}{N(N-1)}\sum_{i<j}{\mathcal{H}(r -||\mathbf{x}(i)-\mathbf{x}(j)||)}
\end{equation*}
with $\mathcal{H}$ the Heaviside step function.
The dimension $n$ then writes :
\begin{equation*}\label{corrdim2}
n = \lim_{r\rightarrow 0}\frac{\log{C(r)}}{\log{r}}
\end{equation*}
In practice, $n$ is approximated by fitting a linear slope of a series of estimates of increasing $r$ and $C$ in logarithmic coordinates. 

As outlined before \cite{Camastra2016}, an ideal  estimator should be robust to noise, high dimensionality and multiscaling, as well as accurate and computationally tractable. Moreover, it should provide a range of values for the input data in which it operates properly. As of today, no single estimator meets all these criteria and using an ensemble of estimators is generally recommended.

Many dimensionality estimators provides a single value for the whole dataset and thus belong to the category of global estimators. Datasets can be composed of complex structures with zones of varying dimensionality. In such a case, the dataset should be explored using local estimators, which estimate ID for each point by looking at its neighbourhood. The neighbourhood is typically defined by taking a ball, with a predetermined fixed radius, centered in the reference points or by considering the $k$ closest neighbours. Such approaches allow repurposing global estimators as local estimators. Notably, it is also possible to partition the data into contiguous areas and compute the dimensionality in each of them. However, this may lead to unwanted border effects. 

The idea behind local ID estimation is to operate at a scale where the manifold can be approximated by its tangent space \cite{Camastra2016}. The data contained in each neighbourhood is thus usually assumed to be uniformly distributed over an $n$-dimensional ball \cite{DANCo,Kerstin,SSV,ANOVA}. In practice, ID proves sensitive to scale and finding an adequate neighbourhood size can be difficult, as it requires finding a trade-off between opposite requirements \cite{Little,Campadelli2015}. Ideally, the neighbourhood should be big relative to the scale of the noise, and contain enough points for the chosen method to work properly. At the same time, it should be small enough to be well approximated by a flat and uniform tangent space.

\section{Estimating intrinsic data dimension based on separability properties}

In the present work, we will follow the framework and notations on estimating the dimensionality of a data point cloud based on the description provided in the works by A.Gorban, I.Tyukin and their colleagues \cite{GORBAN2018303}.

We remind the reader that a point $\mathbf{x}\in R^n$ is linearly separable from a finite set $Y \subset R^n$ if there exists a linear functional $l$ such that $l(\mathbf{x})>l(\mathbf{y})$ for all $y \in Y$. If for any point $\mathbf{x}$ there exists a linear functional separating it from all other data points, then such a data point cloud is called {\it linearly separable} or $1$-convex. The separating functional $l$ may be computed using the linear Support Vector Machine (SVM) algorithms, the Rosenblatt perceptron algorithm, or other comparable methods. However, these computations may be rather costly for large-scale estimates. Hence, it has been suggested to use the simplest non-iterative estimate of the linear functional by Fisher's linear discriminant which is computationally inexpensive, after a well-established standardised pre-processing described below \cite{GORBAN2018303}.

Let us assume that a dataset $X$ is normalized in the following (standardised) way:
\begin{enumerate}
    \item centering
    \item projecting onto the linear subspace spanned by first $k$ principal components, where $k$ may be relatively large
    \item whitening (i.e., applying a linear transformation after which the covariance matrix becomes the identity matrix)
    \item normalising each vector to the unit length, which corresponds to the projection onto a unit sphere.
\end{enumerate}

The 4\textsuperscript{th} transformation (projecting on the sphere) is optional for the general framework previously defined, but it is necessary for comparing the data distribution with a unity sphere. Choosing the number of principal components to retain in the 2\textsuperscript{nd} step of the normalisation has the objective of avoiding excessively small eigenvalues of the covariance matrix (strong collinearity in the data). An effective way to estimate $k$, is by selecting the largest $k$ (in their natural ranking) such that the corresponding eigenvalue $\lambda_k$ is not smaller that $\lambda_1/C$, where $C$ is a predefined threshold. Under most circumstances, $C = 10$ (i.e., the selected eigenvalue is 10 times smaller than the largest one) will result in the most popular linear estimators to work robustly.

After such normalization of $X$, it is said that a point $\mathbf{x} \in X$ is Fisher-linearly separable from the cloud of points $Y$ with parameter $\alpha$, if

\begin{equation*}\label{onepoint_separability_criterion}
(\mathbf{x},\mathbf{y}) \leq \alpha (\mathbf{x},\mathbf{x})
\end{equation*}

\noindent for all $\mathbf{y} \in Y$, where $\alpha \in \left[0,1\right)$. If equation (\ref{onepoint_separability_criterion}) is valid for each point $\mathbf{x} \in X$ such that $Y$ is the set of points $\mathbf{y} \neq \mathbf{x}$ then we call the dataset $X$ Fisher-separable with parameter $\alpha$. In order to quantify deviation from perfect separability, let us introduce $p_\alpha(\mathbf{y})$, a probability that a point $\mathbf{y}$  is separable from all other points. Let us denote $\bar{p}_\alpha(\mathbf{y})$ a mean value of the distribution of $p_\alpha(\mathbf{y})$ over all data points.

Following \cite{GORBAN2018303}, for the equidistribution on the unit sphere $S^{n-1}\in R^n$, $p_\alpha$ does not depend on the data point thanks to the distribution symmetry, and equals:\footnote{In \cite{Gorban2018Unreasonable}, this formula, derived for large $n$, has $\alpha\sqrt{2\pi(n-1)}$ in the denominator. We empirically verified (see Numerical examples section) that changing the denominator to $\alpha\sqrt{2\pi n}$ makes this formula applicable for low dimensions, and the two expressions are very close for large $n$, since $n/(n-1)\rightarrow 1$. In \cite{GORBAN2018303} this formula contains a misprint (personal communication with the authors of \cite{GORBAN2018303}).}.

\begin{equation}\label{UnitSphereDistribution}
p_\alpha=\bar{p}_\alpha=\frac{(1-\alpha^2)^{\frac{n-1}{2}}}{\alpha\sqrt{2\pi n}}
%p_y=\bar{p}_y=\frac{(1-\alpha^2)}{\alpha}
\end{equation}

Therefore, the distribution of $p_\alpha$ for a uniform sampling from an $n-$sphere is a delta function centered in $\bar{p}_\alpha$. The effective dimension of a data set can be evaluated by comparing $\bar{p}_\alpha$ for this data set to the value of $\bar{p}_\alpha$ for the equidistributions on a ball, a sphere, or the Gaussian distribution. Comparison to the sphere is convenient thanks to having an explicit formula (\ref{UnitSphereDistribution}). In order to use this formula, one should project data points on a unit sphere. If $\bar{p}_\alpha$ can be empirically estimated for a given $\alpha$, then the effective dimension can be estimated by solving (\ref{UnitSphereDistribution}) with respect to $n$:

\begin{equation}\label{ComputingEffectiveDimension}
n_\alpha = \frac{W(\frac{-\ln(1-\alpha^2)}{2\pi \bar{p}_\alpha^2\alpha^2(1-\alpha^2)})}{-\ln(1-\alpha^2)}
\end{equation}

\noindent where $W(x)$ is the Lambert function. The self-contained description of the algorithm for computing $n_\alpha$ is provided below (Algorithm~\ref{AlgorithmSepar}).

Based on the above definitions, the fine-grained lumping of the data point cloud can be identified by two interesting features: the histogram of empirical $p_\alpha$ distribution (probabilities of individual point non-separability) and the profile of intrinsic dimensions $n_\alpha$ (\ref{ComputingEffectiveDimension})
for a range of $\alpha$ values (e.g., $\alpha \in [0.6,...,1.0]$). 

\begin{algorithm}
\caption{Computing data point cloud effective dimension from Fisher-separability with parameter $\alpha$}\label{AlgorithmSepar}
\begin{algorithmic}[1]
\STATE $\textit{For a given data matrix } X$
\STATE $\textit{Center the data by columns} X\leftarrow X-\bar{X}$
\STATE $\textit{Apply PCA: } [V,U,S]=PCA(X), $\\$\textit{where }U\textit{ are projections onto principal vectors }V,$ \\$ \textit{and }S\textit{ are explained variances}$
\STATE $\textit{Select the number of components: }$\\ $k = max \{i: S(1)/S(i)<C \}$
\STATE $\textit{For columns of }U, u_i,\textit{ apply data whitening: }$\\ $ u_i \leftarrow u_i/\sigma(u_i), i=1...k$
\STATE $\textit{Project the data vectors onto a unit sphere: }$ \\$u_i \leftarrow u_i/||u_i||, i=1...k$
\STATE $\textit{Compute the Gram matrix } G = UU^T$
\STATE $\textit{Normalize the Gram matrix by the diagonal elements: }$ \\ $G_{ji} \leftarrow G_{ji}/G_{ii}$
\STATE $\textit{Set to zero diagonal elements of G: } G_{ii} = 0$
\STATE $\textit{For each row of G, compute the number of elements}$ \\ $\textit{exceeding $\alpha$: } v_j = \#G_{ji}>\alpha, i,j=1...N$
\STATE $\textit{Compute empirical unseparability probability distribution:}$\\$p_\alpha^j = v_j/N$
\STATE $\textit{Compute empirical mean of $p_\alpha$ :} \bar{p}_\alpha = \frac{1}{N}\sum_{i=1...N}p_\alpha^j$
\STATE $\textit{Compute intrinsic dimension $n_\alpha$ from the formula (\ref{ComputingEffectiveDimension})}$
\end{algorithmic}
\end{algorithm}

\section{Numerical results}

\subsection{Benchmark data}

\begin{figure*}[ht]
{\centering\includegraphics[width=15cm]{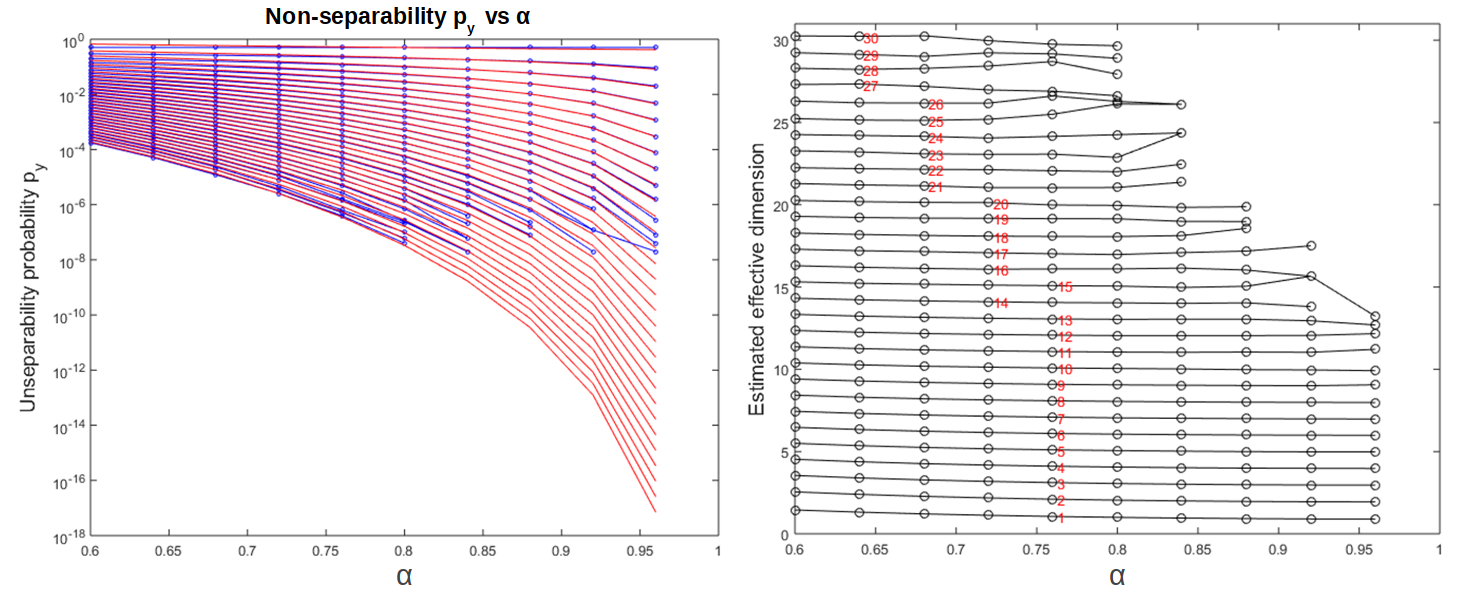}
\caption{Estimating effective dimensions of uniformly sampled unit spheres of various dimensions (from 1 to 30). (Left) Comparison of theoretical (red lines) and empirical (blue lines with markers) estimates of the average unseparability probability $\bar{p}_\alpha$. (Right) Estimated effective dimension of $n$-dimensional spheres ($n=1..30$), as a function of $\alpha$. A single ad-hoc estimate of $n$ is indicated by a number, as such $n_\alpha$ for which $\alpha=0.8\alpha_{max}$, where $\alpha_{max}$ is the maximum value of $\alpha$ for which the empirical $\bar{p}_\alpha>0$.}
\label{UnitSpheres}}.
\end{figure*}

We first checked that the method correctly determines the dimension of uniformly sampled n-dimensional spheres (Figure~\ref{UnitSpheres}). The ability to correctly estimate the dimension in this case depends on the accuracy of estimating the mean empirical unseparability probability for $\alpha$ sufficiently close to 1 which requires a certain number of data points.   

The performance of ID estimation methods is usually assessed on synthetic data consisting of samples generated from $n$-manifolds linearly or non-linearly embedded into a higher dimensional space. The results are then evaluated according to the \textit{mean percentage error}, defined as :
\begin{equation*}
Mean\%error=\frac{100}{\#\{M_i\}}\sum_{i=1}^{\#\{M_i\}}\frac{|\hat{n}_{M_i}-n_{M_i}|}{n_{M_i}}
\end{equation*}
where $\hat{n}_{M_i}$ is the estimated ID and $n_{M_i}$ the true ID of the dataset $M_i$ \cite{Lindheim}.
Different datasets have been used for this purpose. Here, we use the benchmark library made available by Hein and Audibert \cite{Hein}, which is standard across publications as a core benchmark battery. It consists in 13 uniformly sampled manifolds, to which we added isotropic gaussian noise with standard deviation $\sigma=.05$. We also used the ISOMAP Faces dataset \cite{Tenenbaum2319}, which is composed of images from a sculpture's face generated with three degrees of freedom (horizontal pose, vertical pose, lighting direction).  Results are shown for different estimators (\ref{Benchmark}), including those recently published and not covered in the existing reviews \cite{Campadelli2015,Facco2017a,Kerstin}. 

\begin{table*}[ht]
\caption{\label{Benchmark}Predicted ID for synthetic datasets evaluated globally, with added multidimensional isotropic Gaussian noise (standard deviation $\sigma = .05$), and the ISOMAP Faces dataset. Cardinality: Number of points of the dataset, N: embedding dimension, n: intrinsic dimension. \textit{FisherS}: Fisher Separability (The number in parentheses indicates the number of components retained by PCA preprocessing for the separability-based method), \textit{CD}: Correlation Dimension \cite{Grassberger1983}, \textit{GMSTL}: Geodesic Minimum Spanning Tree Length \cite{Costa}, \textit{DANCo}: Dimensionality from Angle and Norm Concentration, \textit{LBMLE}:  Levina-Bickel Maximum Likelihood Estimation \cite{Levina2004}, \textit{ESS}: Expected Simplex Skewness, \textit{FanPCA}: PCA based on \cite{Fan}, \textit{TwoNN}: Two Nearest Neighbors \cite{Facco2017a} }\centering\begin{tabular}{lrrrrrrrrrrr}
\midrule
{} &       Cardinality &       N &     n &  FisherS &     CD &  GMSTL &  DANCo &  LBMLE &    ESS &  FanPCA &  TwoNN \\
\midrule
$\mathbf{M_{13}}$         &  2500&    13&   1&     1.67 (3) &   1.64 &   3.73 &   4&   3.74 &   3.16 &    2&   5.50 \\
$\mathbf{M_{5}}$          &  2500&     3&   2&     2.57 (3) &   2.14 &   2.47 &   3&   2.66 &   2.74 &    1&   2.73 \\
$\mathbf{M_{7}}$          &  2500&     3&   2&     2.94 (3) &   2&   2.24 &   2&   2.39 &   2.93 &    2&   2.67 \\
$\mathbf{M_{11}}$         &  2500&     3&   2&     1.96 (2) &   2.33 &   2.21 &   2&   2.49 &   2.34 &    1&   2.69 \\
$\mathbf{Faces}$        &   698&  4096&   3&    3.12 (28) &   0.78 &   1.64 &   4&   4.31 &   7.49 &    8&   3.49 \\
$\mathbf{M_{2}}$          &  2500&     5&   3&     2.66 (3) &   3.60 &   4.61 &   4&   4.42 &   2.66 &    2&   4.69 \\
$\mathbf{M_{3}}$          &  2500&     6&   4&     2.87 (4) &   3.16 &   3.36 &   4&   4.40 &   3.11 &    2&   4.36 \\
$\mathbf{M_{4}}$          &  2500&     8&   4&     5.78 (8) &   3.90 &   4.33 &   4&   4.38 &   7.79 &    5&   3.96 \\
$\mathbf{M_{6}}$          &  2500&    36&   6&     8.50 (12) &   5.99 &   6.62 &   7&   7.05 &  11.98 &    9&   6.27 \\
$\mathbf{M_{1}}$          &  2500&    11&  10&    11.03 (11) &   8.96 &   9.02 &  11&   9.88 &  10.81 &    7&   9.43 \\
$\mathbf{M_{10a}}$      &  2500&    11&  10&     9.46 (10) &   7.86 &   9.50 &  10&   8.90 &  10.31 &    7&   8.57 \\
$\mathbf{M_{8}}$     &  2500&    72&  12&    17.41 (24) &  10.97 &  13.04 &  17&  14.74 &  24.11 &   18&  13.15 \\
$\mathbf{M_{10b}}$        &  2500&    18&  17&    15.94 (17) &  11.88 &  13.15 &  16&  13.89 &  17.35 &   13&  13.59 \\
$\mathbf{M_{12}}$         &  2500&    20&  20&    19.83 (20) &  10.62 &  16.05 &  20&  17.07 &  19.90 &   11&  16.94 \\
$\mathbf{M_{9}}$          &  2500&    20&  20&    19.07 (20) &  13.51 &  14.26 &  19&  15.73 &  20.26 &   11&  15.68 \\
$\mathbf{M_{10c}}$      &  2500&    25&  24&    22.62 (24) &  15.15 &  21.94 &  23&  18.24 &  24.42 &   17&  17.36 \\
$\mathbf{M_{10d}}$       &  2500&    71&  70&    68.74 (70) &  29.89 &  36.62 &  71&  38.92 &  71.95 &   43&  39.18 \\
\midrule
$\mathbf{Mean \% error}$ &      &      &    &    28.82 &  32.45 &  36.35 &  43.04 &  43.83 &  66.78 &   67.56 &  74.91 \\
\midrule
\end{tabular}\end{table*}

We find that Fisher separability is an accurate estimator of ID across the manifold library. Notably, it is one of the few methods performing well in high dimension. Indeed, methods exploiting concentration of measure (\textit{FisherS}, \textit{DANCo}, \textit{ESS}) manage to give a close estimate for $\mathbf{M_{10d}}$, a 70-cube, while all other methods largely underestimate the dimension. We observed that the performance of the Fisher separability method was close with \textit{DANCo} and \textit{ESS}. However, \textit{FisherS} estimated well small effective dimensions in addition to large ones. Both \textit{ESS} (implemented in R) and \textit{FisherS} (implemented in Python 3) are faster than \textit{DANCo} (implemented in MATLAB), which scales worse with respect to increasing dimension (respectively 500ms, 1.9s, 25.6s on $\mathbf{M_{10d}}$, over an average of 7 runs).

\begin{figure*}[!htb]
{\centering\includegraphics[width=17cm]{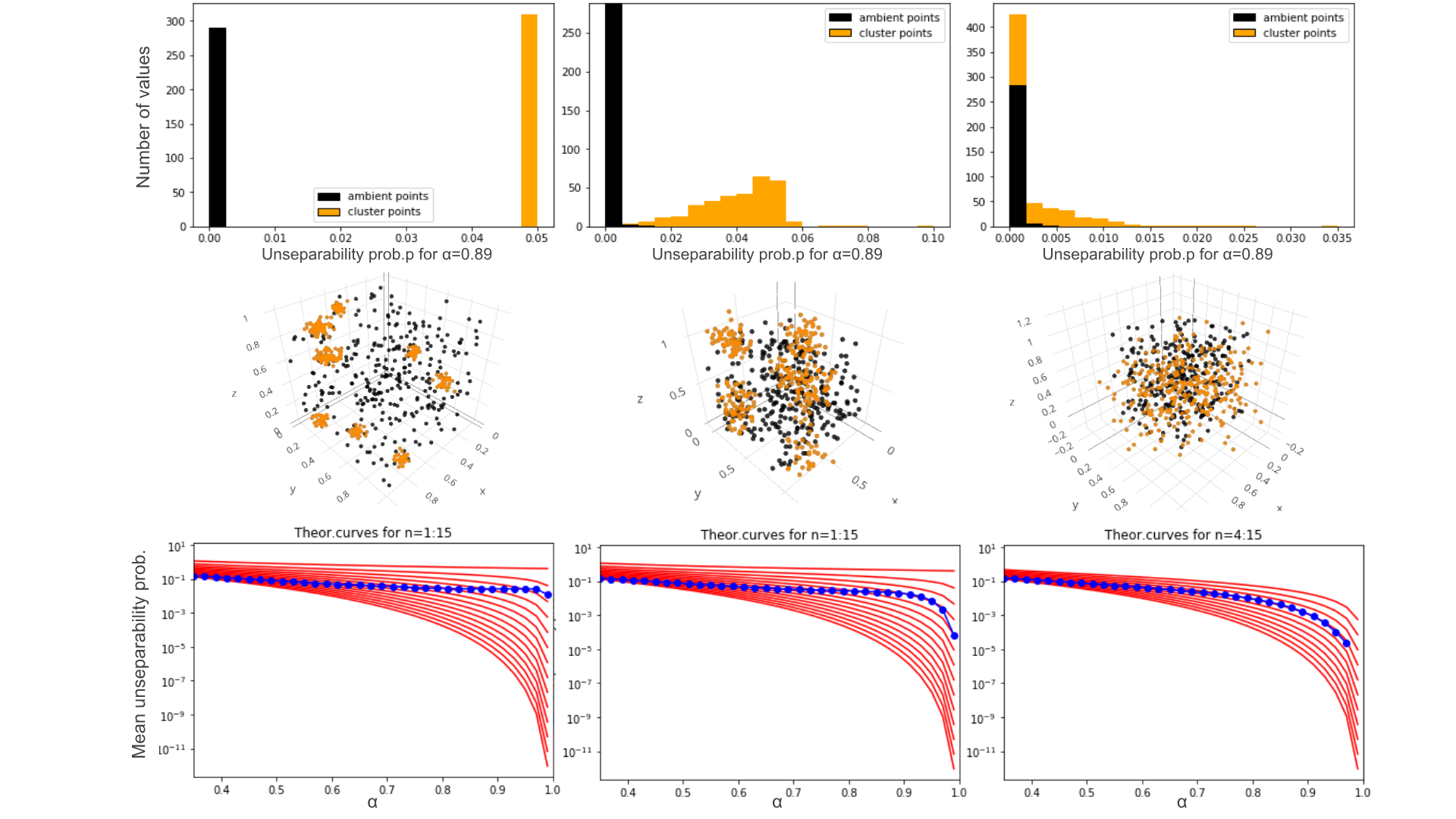}}
\caption{Illustration of the presented method on datasets consisting of samples from a uniform distribution in the unity 10-dimensional cube, and of 10 clusters formed by choosing random center points to sample uniform 10-balls with radiuses (from left to right) $0.1$, $0.3$, and $0.6$. First row: histogram of unseparability probability. Second row: 3D scatter plot of the datasets. Third row: The empirical mean unseparability probability as a function of parameter $\alpha$ value (blue) shown on top of the theoretical curves (red) as the clusters become fuzzier due to increasing radius.
\label{Microclusters}}.
\end{figure*}

Additionally, we generated three versions of a dataset with random clusters, to illustrate the idea of characterising the effect of having non-homogeneity in the data cloud on the separability probability distribution. These datasets consist in a mixture of samples from a uniform distribution and from uniformly sampled balls centered at random points (\ref{Microclusters}).

The complete analysis containing more detailed results is available as an interactive Python 3 notebook, including the necessary code to test additional methods and manifolds. The notebook can interface methods in various languages and thus be a useful basis to perform future benchmark tests.

\subsection{Cancer somatic mutation data: an example of fine-grained lumping}

Cancer a complex disease, that is largely caused by the accumulation of somatic mutation during the lifetime of cells in the organism body. Large-scale genomic profiling provides information on which genes are mutated in the cells composing a tumor at the moment of cancer diagnosis and there is a hope that this information can help driving therapeutic decisions. However, application of standard machine learning methods for this kind of data is difficult because of their extreme sparsity and non-homogeneity of mutation profiles\cite{LeMorvan2017}. Mutation matrix (genes vs tumor) in its  simplest form is a binary matrix marking non-sense or missense mutation of a certain gene in a chohort of tumors. Because there exists very small overlap between mutation profiles in any two tumors, the data cloud representing a mutation matrix is usually thought to be high-dimensional and suffering from the curse of dimensionality.

\begin{figure*}[!htb]
{\centering \includegraphics[width=17cm]{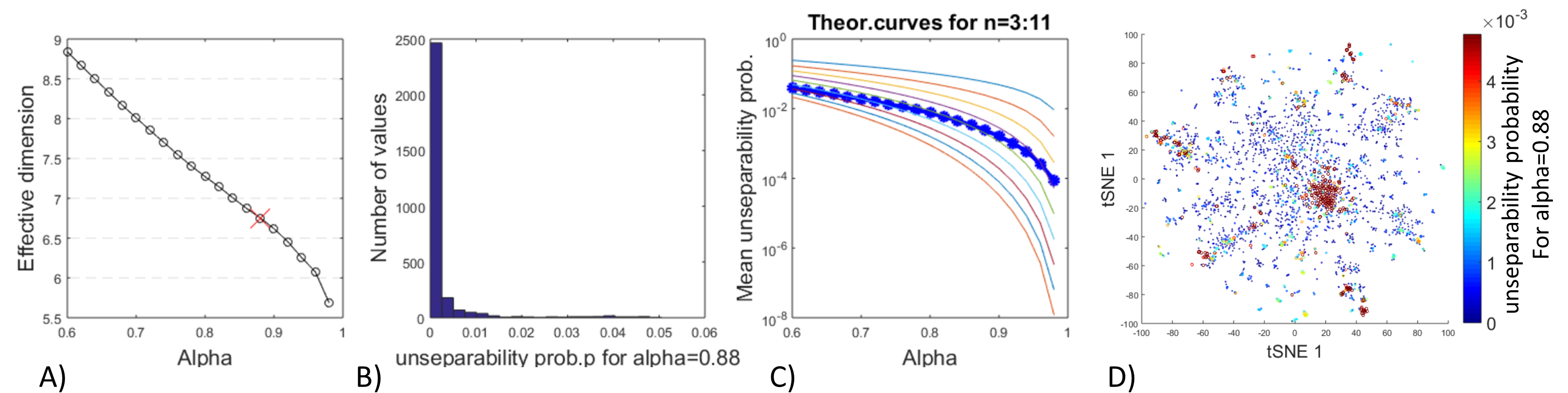}}
\caption{Analysis of breast cancer somatic mutation data. The initial dataset represents a binary matrix (genes vs tumors) with ones marking any deleterious mutation in a gene found in a tumor. A) Plot showing a range of estimated effective dimensions for a range of $\alpha$. An single ad-hoc estimate is shown by cross. B) Distribution of unseparability probability distribution for a particular value of $\alpha=0.88$. C) Empirical estimates for the mean unseparability probability for several $\alpha$ values, shown on top of the theoretical curves for n-dimensional uniformly sampled spheres (starting from n=3). D) tSNE visualization of the dataset (each point corresponds to a gene). Colors show the estimated empirical unseparabity probability $p_\alpha$ for a given data point.
\label{mutation_data}}.
\end{figure*}

We obtained the mutation matrix for 945 breast cancer tumors from The Genome Cancer Atlas (TCGA) as it is provided in \cite{LeMorvan2017}. After filtering genes having less than 5 mutations in all tumors, we were left with 2932 genes. For each tumor, we divided its binary mutation profile by the total number of mutations in this tumor, in order to compensate for large differences in total mutational load between tumors. We analyzed the data point cloud where each point corresponded to a gene, and studied its separability properties using Algorithm~\ref{AlgorithmSepar}. The criterion used from Algorithm~\ref{AlgorithmSepar} for determining the number of principal components selected 34 dimensions, indicating relatively large dimension of the linear manifold approximating the mutation data. Despite this, the separability analysis showed that the separability properties of this data cloud is close to the uniformly sampled $7$-dimensional sphere (Figure~\ref{mutation_data},A,C).

We observed that the $p_\alpha$ probability distributions were overall close to the delta function (Figure~\ref{mutation_data}B), indicating good separability properties of the data cloud. However, there was a non-negligible fraction of data points which could not be separated from the rest of the data cloud even for relatively large $\alpha=0.88$. We further visualized the data point cloud by applying t-distributed stochastic neighbour embedding (t-SNE)\cite{vanDerMaaten2008}, which showed existence of small clusters where the points are less separable, embedded into the sparse cloud of separable points.

\subsection{Highlighting the variable complexity of single cell datasets through separability analysis}

Single cell transcriptomics allows the simultaneous measurement of thousands of genes across tens of thousands of cells, resulting in potentially very complex biological big data that can be used to identify cell types or even reconstruct the dynamics of biological differentiation \cite{elpigraph, stream}.

In a recent work, this technology has been used to explore the different cell types contained in an adult organism of the regenerative planarian Schmidtea mediterranea \cite{PlanarianSC}. Using these data, the authors has been able to identify (via computational analysis) 51 different cell types and the transcriptional changes associated with the commitment of different stem cells (\textit{neoblasts}) into various subpopulations.

\begin{figure*}[ht]
{\centering\includegraphics[width=18cm]{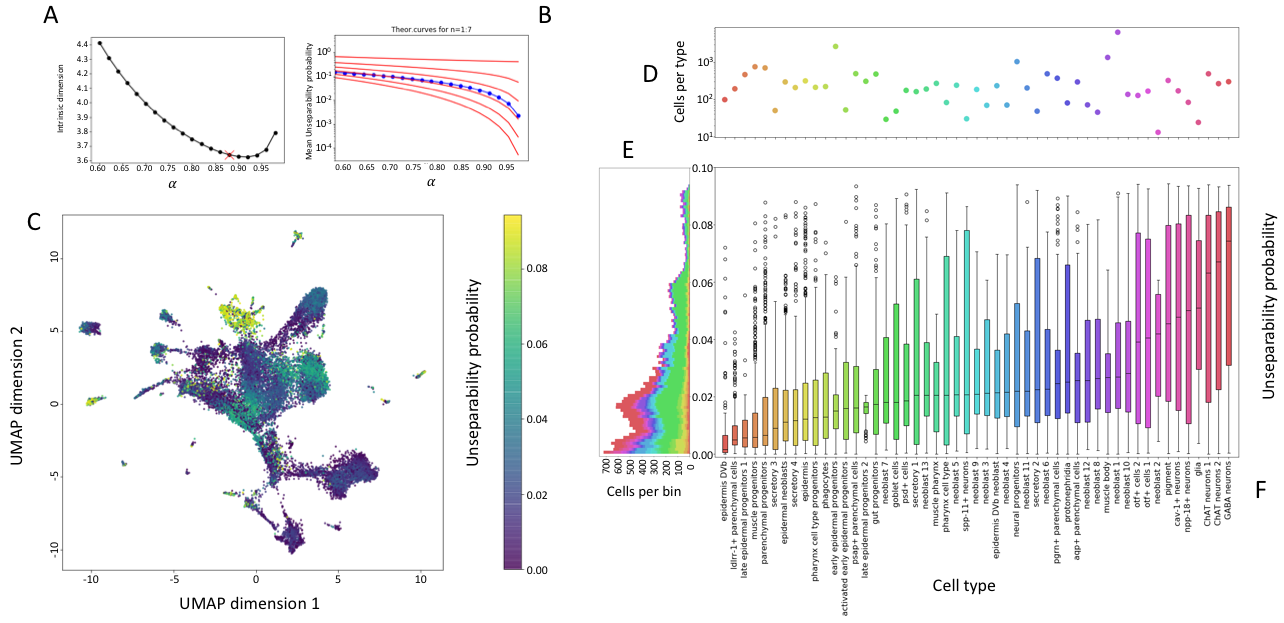}}
\caption{ID of the planarian transcriptome with data from \cite{PlanarianSC}. \textbf{(A-B)} ID  and mean unseparability probability as a function of $\alpha$ with the same graphic conventions used before. \textbf{(C)} UMAP projection in 2D of the cells of the datasets color-coded according to their unseparability probaiblity. \textbf{(D)} Number of cells of different types. Note the logarithmic scale on the y axis. \textbf{(E)} Histogram of unseparability probability for $\alpha = 0.88$. \textbf{(F)} Boxplot of unseparability probability for $\alpha = 0.88$ by cell type. Note that in panels D-E the different colours indicate the different cell types identified in the original publication.
\label{Plan}}.
\end{figure*}

Given the complex nature of the data, we decided to use Fisher separability to highlight  potential biological properties. After a standard preprocessing pipeline, which included selection of the overdispersed genes and log-transformation of gene expression, the datasets contained 21612 cells characterised by 4515 genes. After an initial filtering that retained 7 PCs, our analysis estimates a global ID close to 4 (\ref{Plan}A). By looking at the unseparability probability per cell, we can further appreciate how separability varies across different parts of the dataset (\ref{Plan}C). To further explore this aspect we looked at the distribution of unseparability probability per cell type.

Interestingly, different populations show a tendency to have different ranges of unseparability probabilities (\ref{Plan}E-F) which cannot be explained by the population size (\ref{Plan}D). The presence of a multi-peak distribution (\ref{Plan}E) indicates the presence of multiple \textit{dimensionality scales} and suggests the presence of micro/meso-clusters embedded into a more uniform manifold. 

Remarkably, neurons tend to have a larger unseparability probability, an indication of locally compact distribution and hence of a potentially structured heterogeneity, while epidermal cells are on the other end of the spectrum. The different neoblast populations display a comparable range of unseparability probability, which sits somewhere in the middle, a potential indication of a controlled heterogeneity.

\section{Implementation}

We provide MATLAB and Python 3 implementations of ID estimation based on data point cloud Fisher-separability at \url{https://github.com/auranic/FisherSeparabilityAnalysis}. We also provide a Python 3 notebook for performing the calculations of the global and local effective dimensionality, using various estimators.

\section{Conclusion}

In this paper, we have exploited the framework of linear Fisher separability in order to estimate the intrinsic dimension of both synthetic and real-life biological datasets. The suggested approach does not assume the presence of a low-dimensional variety around which the data point cloud is organized. According to this framework, deviations from uniformity of data sampling lead to a decrease in the intrinsic dimensionality. Despite this general assumption, the approach demonstrated a surprisingly good performance even for estimating the dimensionality of datasets representing noisy samples from embedded manifolds. The advantages of the method manifest in its efficiency across a wide range of dimensions, robustness to noise, and ability to quantify the presence of fine-grained lumping in the data.

Structures found in the data point clouds resulting from applications of modern biotechnologies might reflect details of molecular mechanisms shaping life. Indeed, computational biology approaches have been capable to gain new insights from mining large-scale molecular datasets and to provide new information that continuously improve our understanding of life and suggest new therapeutic avenues to treat diseases such as cancer. In this paper, we demonstrate how the suggested approach can be used in exploring the structure of two data types that are generally considered to be hard to analyse  (mutation and single cell RNA-Seq data) and concluded that separability analysis can provide insights into the organization of their data point clouds.

%\section*{Acknowledgment}

%The project is supported by the Ministry of education and science of Russia (Project No. 14.Y26.31.0022)
%and by Big Data PSL Research University project "PSL Institute for Data Science" (2016--2018).

%\section*{References}
\bibliographystyle{IEEEtran}
\bibliography{Separability}

\begin{thebibliography}{10}
\providecommand{\url}[1]{#1}
\csname url@samestyle\endcsname
\providecommand{\newblock}{\relax}
\providecommand{\bibinfo}[2]{#2}
\providecommand{\BIBentrySTDinterwordspacing}{\spaceskip=0pt\relax}
\providecommand{\BIBentryALTinterwordstretchfactor}{4}
\providecommand{\BIBentryALTinterwordspacing}{\spaceskip=\fontdimen2\font plus
\BIBentryALTinterwordstretchfactor\fontdimen3\font minus
  \fontdimen4\font\relax}
\providecommand{\BIBforeignlanguage}[2]{{%
\expandafter\ifx\csname l@#1\endcsname\relax
\typeout{** WARNING: IEEEtran.bst: No hyphenation pattern has been}%
\typeout{** loaded for the language `#1'. Using the pattern for}%
\typeout{** the default language instead.}%
\else
\language=\csname l@#1\endcsname
\fi
#2}}
\providecommand{\BIBdecl}{\relax}
\BIBdecl

\bibitem{Campadelli2015}
\BIBentryALTinterwordspacing
P.~Campadelli, E.~Casiraghi, C.~Ceruti, and A.~Rozza, ``{Intrinsic Dimension
  Estimation: Relevant Techniques and a Benchmark Framework},''
  \emph{Mathematical Problems in Engineering}, vol. 2015, pp. 1--21, 2015.
  [Online]. Available: \url{http://www.hindawi.com/journals/mpe/2015/759567/}
\BIBentrySTDinterwordspacing

\bibitem{Camastra2016}
\BIBentryALTinterwordspacing
F.~Camastra and A.~Staiano, ``{Intrinsic dimension estimation: Advances and
  open problems},'' \emph{Information Sciences}, vol. 328, pp. 26--41, jan
  2016. [Online]. Available:
  \url{https://www.sciencedirect.com/science/article/pii/S0020025515006179}
\BIBentrySTDinterwordspacing

\bibitem{GORBAN2007382}
\BIBentryALTinterwordspacing
A.~Gorban, N.~Sumner, and A.~Zinovyev, ``Topological grammars for data
  approximation,'' \emph{Applied Mathematics Letters}, vol.~20, no.~4, pp. 382
  -- 386, 2007. [Online]. Available:
  \url{http://www.sciencedirect.com/science/article/pii/S0893965906001856}
\BIBentrySTDinterwordspacing

\bibitem{Gorban2009}
A.~N. Gorban and A.~Zinovyev, ``Principal graphs and manifolds,'' \emph{In
  Handbook of Research on Machine Learning Applications and Trends: Algorithms,
  Methods and Techniques, eds. Olivas E.S., Guererro J.D.M., Sober M.M.,
  Benedito J.R.M., Lopes A.J.S.}, 2009.

\bibitem{elpigraph}
\BIBentryALTinterwordspacing
L.~Albergante, E.~M. Mirkes, H.~Chen, A.~Martin, L.~Faure, E.~Barillot,
  L.~Pinello, A.~N. Gorban, and A.~Zinovyev, ``Robust and scalable learning of
  data manifolds with complex topologies via elpigraph,'' \emph{CoRR}, vol.
  abs/1804.07580, 2018. [Online]. Available:
  \url{http://arxiv.org/abs/1804.07580}
\BIBentrySTDinterwordspacing

\bibitem{Gorban2018Blessing}
A.~N. Gorban and I.~Y. Tyukin, ``{{B}lessing of dimensionality: mathematical
  foundations of the statistical physics of data},'' \emph{Philos Trans A Math
  Phys Eng Sci}, vol. 376, no. 2118, Apr 2018.

\bibitem{GORBAN2018303}
\BIBentryALTinterwordspacing
A.~Gorban, A.~Golubkov, B.~Grechuk, E.~Mirkes, and I.~Tyukin, ``Correction of
  ai systems by linear discriminants: Probabilistic foundations,''
  \emph{Information Sciences}, vol. 466, pp. 303 -- 322, 2018. [Online].
  Available:
  \url{http://www.sciencedirect.com/science/article/pii/S0020025518305607}
\BIBentrySTDinterwordspacing

\bibitem{Gorban2017Theorems}
A.~N. Gorban and I.~Y. Tyukin, ``{{S}tochastic separation theorems},''
  \emph{Neural Netw}, vol.~94, pp. 255--259, Oct 2017.

\bibitem{Tykin2018Front}
I.~Y. Tyukin, A.~N. Gorban, K.~I. Sofeykov, and I.~Romanenko, ``{{K}nowledge
  {T}ransfer {B}etween {A}rtificial {I}ntelligence {S}ystems},'' \emph{Front
  Neurorobot}, vol.~12, p.~49, 2018.

\bibitem{Bennett}
R.~Bennett, ``The intrinsic dimensionality of signal collections,'' \emph{IEEE
  Transactions on Information Theory}, vol.~15, no.~5, pp. 517--525, September
  1969.

\bibitem{bishop1995neural}
C.~M. Bishop \emph{et~al.}, \emph{Neural networks for pattern
  recognition}.\hskip 1em plus 0.5em minus 0.4em\relax Oxford university press,
  1995.

\bibitem{Fuku1982}
\BIBentryALTinterwordspacing
K.~Fukunaga, \emph{Intrinsic dimensionality extraction}, ser. in: P.R.
  Krishnaiah, L.N. Kanal (Eds.), Pattern Recognition and Reduction of
  Dimensionality, Handbook of Statistics, Vol. 2, North-Holland, Amsterdam,
  1982, pp. 347–362, 1982. [Online]. Available:
  \url{https://doi.org/10.1016/S0169-7161(82)02018-5}
\BIBentrySTDinterwordspacing

\bibitem{Kerstin}
K.~Johnsson, ``\BIBforeignlanguage{English}{Structures in high-dimensional
  data: Intrinsic dimension and cluster analysis},'' Ph.D. dissertation,
  Faculty of Engineering, LTH, 8 2016.

\bibitem{mordohai2010}
P.~Mordohai and G.~Medioni, ``Dimensionality estimation, manifold learning and
  function approximation using tensor voting,'' \emph{Journal of Machine
  Learning Research}, vol.~11, no. Jan, pp. 411--450, 2010.

\bibitem{li2009}
C.-G. Li, J.~Guo, and B.~Xiao, ``Intrinsic dimensionality estimation within
  neighborhood convex hull,'' \emph{International Journal of Pattern
  Recognition and Artificial Intelligence}, vol.~23, no.~01, pp. 31--44, 2009.

\bibitem{shepard1972multidimensional}
R.~N. Shepard, A.~K. Romney, and S.~B. Nerlove, ``Multidimensional scaling:
  Theory and applications in the behavioral sciences: Vol.: 1: Theory.''\hskip
  1em plus 0.5em minus 0.4em\relax Seminar Press New York, 1972.

\bibitem{jolliffe2002principal}
\BIBentryALTinterwordspacing
I.~Jolliffe, \emph{Principal Component Analysis}, ser. Springer Series in
  Statistics.\hskip 1em plus 0.5em minus 0.4em\relax Springer, 2002. [Online].
  Available: \url{https://books.google.fr/books?id=\_olByCrhjwIC}
\BIBentrySTDinterwordspacing

\bibitem{Tenenbaum2319}
\BIBentryALTinterwordspacing
J.~B. Tenenbaum, V.~d. Silva, and J.~C. Langford, ``A global geometric
  framework for nonlinear dimensionality reduction,'' \emph{Science}, vol. 290,
  no. 5500, pp. 2319--2323, 2000. [Online]. Available:
  \url{http://science.sciencemag.org/content/290/5500/2319}
\BIBentrySTDinterwordspacing

\bibitem{Costa}
J.~A. Costa and A.~O. Hero, ``Geodesic entropic graphs for dimension and
  entropy estimation in manifold learning,'' \emph{IEEE Transactions on Signal
  Processing}, vol.~52, no.~8, pp. 2210--2221, August 2004.

\bibitem{DANCo}
\BIBentryALTinterwordspacing
C.~Ceruti, S.~Bassis, A.~Rozza, G.~Lombardi, E.~Casiraghi, and P.~Campadelli,
  ``{DANCo: An intrinsic dimensionality estimator exploiting angle and norm
  concentration},'' \emph{Pattern Recognition}, vol.~47, no.~8, pp. 2569--2581,
  aug 2014. [Online]. Available:
  \url{https://www.sciencedirect.com/science/article/pii/S003132031400065X}
\BIBentrySTDinterwordspacing

\bibitem{ANOVA}
M.~Díaz, A.~J. Quiroz, and M.~Velasco, ``Local angles and dimension estimation
  from data on manifolds,'' 2018.

\bibitem{SSV}
\BIBentryALTinterwordspacing
D.~R. Wissel, ``Intrinsic dimension estimation using simplex volumes,'' Ph.D.
  dissertation, 2018. [Online]. Available:
  \url{http://hss.ulb.uni-bonn.de/2018/4951/4951.htm}
\BIBentrySTDinterwordspacing

\bibitem{Grassberger1983}
\BIBentryALTinterwordspacing
P.~Grassberger and I.~Procaccia, ``{Measuring the strangeness of strange
  attractors},'' \emph{Physica D: Nonlinear Phenomena}, vol.~9, no. 1-2, pp.
  189--208, oct 1983. [Online]. Available:
  \url{https://www.sciencedirect.com/science/article/pii/0167278983902981}
\BIBentrySTDinterwordspacing

\bibitem{Little}
\BIBentryALTinterwordspacing
A.~V. Little, Y.-M. Jung, and M.~Maggioni, ``{Multiscale Estimation of
  Intrinsic Dimensionality of Data Sets},'' Tech. Rep., 2009. [Online].
  Available: \url{www.aaai.org}
\BIBentrySTDinterwordspacing

\bibitem{Gorban2018Unreasonable}
\BIBentryALTinterwordspacing
A.~N. Gorban, V.~A. Makarov, and I.~Y. Tyukin, ``The unreasonable effectiveness
  of small neural ensembles in high-dimensional brain,'' \emph{Physics of Life
  Reviews}, Oct 2018. [Online]. Available:
  \url{http://dx.doi.org/10.1016/j.plrev.2018.09.005}
\BIBentrySTDinterwordspacing

\bibitem{Lindheim}
\BIBentryALTinterwordspacing
J.~V. Lindheim, ``On intrinsic dimension estimation and minimal diffusion maps
  embeddings of point clouds,'' Master's thesis, Freien Universität Berlin,
  2018. [Online]. Available:
  \url{http://www.zib.de/ext-data/manifold-learning/thesis.pdf}
\BIBentrySTDinterwordspacing

\bibitem{Hein}
M.~Hein and J.-Y. Audibert, ``Intrinsic dimensionality estimation of
  submanifolds in r d,'' in \emph{Proceedings of the 22nd international
  conference on Machine learning}.\hskip 1em plus 0.5em minus 0.4em\relax ACM,
  2005, pp. 289--296.

\bibitem{Facco2017a}
\BIBentryALTinterwordspacing
E.~Facco, M.~D'Errico, A.~Rodriguez, and A.~Laio, ``{Estimating the intrinsic
  dimension of datasets by a minimal neighborhood information},''
  \emph{Scientific Reports}, vol.~7, no.~1, p. 12140, dec 2017. [Online].
  Available: \url{http://www.nature.com/articles/s41598-017-11873-y}
\BIBentrySTDinterwordspacing

\bibitem{Levina2004}
\BIBentryALTinterwordspacing
E.~Levina and P.~J. Bickel, ``{Maximum Likelihood estimation of intrinsic
  dimension},'' pp. 777--784, 2004. [Online]. Available:
  \url{https://dl.acm.org/citation.cfm?id=2976138}
\BIBentrySTDinterwordspacing

\bibitem{Fan}
\BIBentryALTinterwordspacing
M.~Fan, N.~Gu, H.~Qiao, and B.~Zhang, ``{Intrinsic dimension estimation of data
  by principal component analysis},'' Tech. Rep., 2010. [Online]. Available:
  \url{https://pdfs.semanticscholar.org/7d4d/936df2f628550123cef40996277628468e61.pdf}
\BIBentrySTDinterwordspacing

\bibitem{LeMorvan2017}
M.~Le~Morvan, A.~Zinovyev, and J.~P. Vert, ``{{N}et{N}or{M}: {C}apturing
  cancer-relevant information in somatic exome mutation data with gene networks
  for cancer stratification and prognosis},'' \emph{PLoS Comput. Biol.},
  vol.~13, no.~6, p. e1005573, Jun 2017.

\bibitem{vanDerMaaten2008}
\BIBentryALTinterwordspacing
L.~van~der Maaten and G.~Hinton, ``Visualizing data using {t-SNE},''
  \emph{Journal of Machine Learning Research}, vol.~9, pp. 2579--2605, 2008.
  [Online]. Available: \url{http://www.jmlr.org/papers/v9/vandermaaten08a.html}
\BIBentrySTDinterwordspacing

\bibitem{stream}
\BIBentryALTinterwordspacing
H.~Chen, L.~Albergante, J.~Y. Hsu, C.~A. Lareau, G.~Lo~Bosco, J.~Guan, S.~Zhou,
  A.~N. Gorban, D.~E. Bauer, M.~J. Aryee, D.~M. Langenau, A.~Zinovyev, J.~D.
  Buenrostro, G.-C. Yuan, and L.~Pinello, ``Stream: Single-cell trajectories
  reconstruction, exploration and mapping of omics data,'' \emph{bioRxiv},
  2018. [Online]. Available:
  \url{https://www.biorxiv.org/content/early/2018/04/18/302554.1}
\BIBentrySTDinterwordspacing

\bibitem{PlanarianSC}
\BIBentryALTinterwordspacing
M.~Plass, J.~Solana, F.~A. Wolf, S.~Ayoub, A.~Misios, P.~Gla{\v z}ar,
  B.~Obermayer, F.~J. Theis, C.~Kocks, and N.~Rajewsky, ``Cell type atlas and
  lineage tree of a whole complex animal by single-cell transcriptomics,''
  \emph{Science}, vol. 360, no. 6391, 2018. [Online]. Available:
  \url{http://science.sciencemag.org/content/360/6391/eaaq1723}
\BIBentrySTDinterwordspacing

\end{thebibliography}

\end{document}